\documentclass{article}
\usepackage{iclr2027_conference,times}
\usepackage{amsmath,amssymb}
\usepackage{algorithm}
\usepackage{algpseudocode}
\usepackage{graphicx,booktabs,array,xcolor}
\usepackage{microtype}
\usepackage{hyperref}
\usepackage{url}
\hypersetup{hidelinks,pdfborder={0 0 0},pdfauthor={},pdftitle={From Agents to Edges: Paired Optimization for Multi-Agent Collaboration}}
\newcommand{\method}{PACO}

\title{From Agents to Edges: Paired Optimization for Multi-Agent Collaboration}
\author{
\parbox{\dimexpr\textwidth-2\tabcolsep\relax}{
  \centering
  Feng Lin, ~~Xiaotian Yu, ~~Yifan Yang, ~~Aijun Li\\
  \textnormal{Intellifusion Inc.}
}
}

\begin{document}
\iclrfinalcopy.
\maketitle
\lhead{Preprint}
\begin{abstract}

Strong individual agents do not guarantee effective collaboration. Communication must align what one agent conveys with how another interprets and uses it. This dependency makes the communication edge a natural unit of optimization. We introduce Paired Automated Communication Optimization (PACO), which jointly optimizes the instructions at both ends of each edge and their shared communication convention in systems built on black-box language models. PACO keeps model weights, agent roles, and workflow topology fixed. A state-aware discounted bandit selects edges for refinement, while execution feedback guides paired revisions evaluated through staged end-to-end execution. The optimized instructions integrate into existing agent prompts without additional model calls at deployment. On a four-agent workflow evaluated across diverse multi-agent benchmarks, PACO consistently outperforms leading prompt-optimization and communication baselines. Configurations learned on HotpotQA retain their advantage in cross-dataset transfer without target-specific optimization. These results demonstrate that optimizing how agents communicate can substantially improve collaboration within an otherwise fixed system.
\end{abstract}

\section{Introduction}
\label{sec:intro}

Multi-agent systems (MAS) built on large language models (LLMs) distribute complex tasks across specialized agents, with applications in software development \citep{qian2024chatdev}, GPU kernel optimization \citep{luo2026agentic}, web automation \citep{fourney2024magentic}, and scientific discovery \citep{swanson2025virtuallab}. Their effectiveness depends on more than the quality of individual agents: each handoff determines the context available for subsequent reasoning. When agents have different information access or responsibilities, a correct intermediate result may still be insufficient if its evidence, constraints, or uncertainty are omitted or misinterpreted downstream \citep{cemri2025multi}. Such failures can undermine the benefit of specialization even when the underlying models are capable. We study how to improve these handoffs within a fixed workflow by optimizing the instructions that govern context construction and use.

To illustrate this problem, consider a simple distributed averaging task, following the classical distinction between local aggregates and the intermediate information needed to combine them \citep{gray1997datacube}. An upstream agent observes two values, $10$ and $10$, and correctly reports a local mean of $10$. A downstream agent holds another value, $100$, and must compute the mean across all three observations. If the message contains only the local mean, the downstream agent lacks the sample count needed to weight it correctly. Even if that count is provided, averaging the two means without weighting yields $55$ instead of the correct result, $40$. 
The relevant dependency spans two agents: the upstream agent must provide the information required by the aggregation rule, and the downstream agent must apply that rule consistently. This motivates grouping the instructions governing a handoff into one optimization unit.

\begin{figure}[t]
    \centering
    \includegraphics[width=0.92\linewidth]{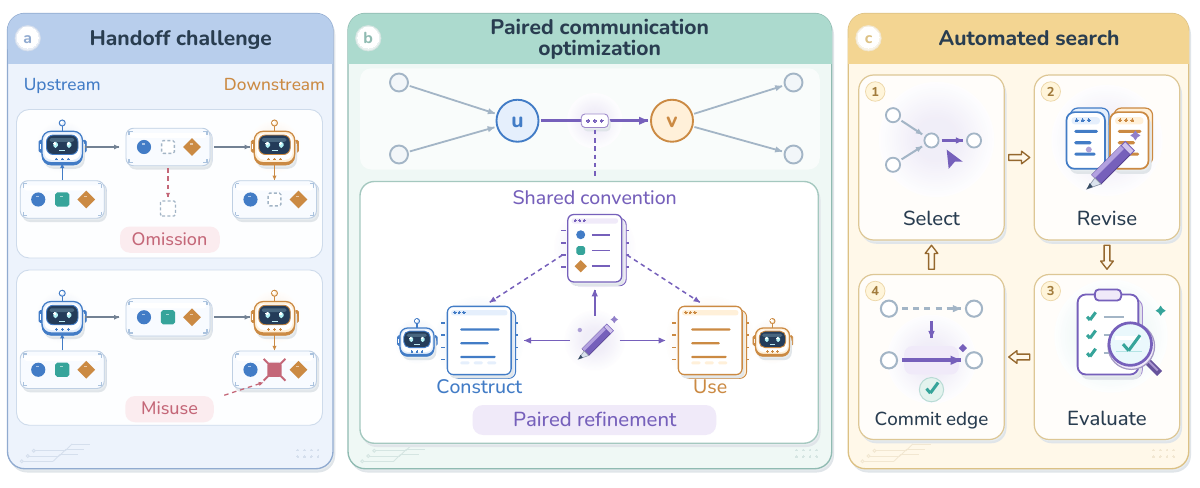}
    \vspace{-5pt}
    \caption{\textbf{Overview of PACO.} (a) Task context can be omitted during handoff or misused downstream. (b) PACO jointly optimizes upstream and downstream instructions
    together with their shared communication convention. (c) An automated loop selects edges, refines paired instructions, evaluates candidates, and commits accepted revisions.}
    \vspace{-15pt}
    \label{fig:paco-overview}
\end{figure}

Existing approaches improve collaboration through workflow search \citep{aflow2025,maas2025}, communication protocols and message representations \citep{huang2026pact,gupta2026optimacs}, and agent or component prompt optimization \citep{gepa2026,wang2026maspo,hong2026maspob}. Several already account explicitly for interactions: MAPRO models compatibility between agent prompts \citep{zhang2026mapro}, MASPOB supports block-coordinate updates of adjacent agents \citep{hong2026maspob}, and AgentAsk introduces edge-level clarification during execution \citep{lin2026agentask}. These advances motivate a more specific question: how should communication-dependent instructions be grouped for optimization? We organize them around a directed edge, combining the upstream agent's outgoing instructions, the downstream agent's incoming instructions, and their shared communication convention into one editable package. This parameterization makes the handoff itself an explicit search variable while keeping role instructions and other communication blocks fixed during an ordinary edge update.

We introduce \textbf{P}aired \textbf{A}utomated
\textbf{C}ommunication \textbf{O}ptimization (PACO), an edge-centric
framework that automates communication-instruction refinement
in multi-agent systems built on black-box LLMs.
As illustrated in Figure~\ref{fig:paco-overview}, PACO jointly
refines the upstream instructions for constructing task context,
the downstream instructions for interpreting and using it,
and their shared communication convention.
A state-aware discounted bandit selects edges using
configuration-specific search histories. Execution feedback
guides two rounds of paired revision, allowing proposals to be
refined after their effects are observed. Staged end-to-end
evaluations screen candidates on small batches before evaluating
selected candidates on the internal validation set.
Throughout optimization, model weights, agent roles, topology,
and information access remain fixed.
The learned packages are incorporated into existing agent prompts,
so deployment requires no additional message-rewriting or
clarification calls.

Experiments on a four-agent workflow across three question-answering
benchmarks \citep{yang-etal-2018-hotpotqa,dua2019drop,trivedi2022musique}
demonstrate the value of optimizing communication.
PACO consistently outperforms leading prompt-optimization and
communication baselines, improving HotpotQA Joint F1 by over
30 points relative to the unoptimized workflow.
Configurations learned on HotpotQA retain their advantage under
cross-dataset transfer without target-specific optimization,
with a gain of nearly 6 Answer F1 points over GEPA on DROP,
the strongest evaluated baseline.
Controlled ablations support the value of paired refinement
and assess the search components, while execution analyses
characterize changes in reference preservation and downstream
interpretation.

Our main contributions are:
\begin{itemize}
    \item \textbf{From agents to edges.}
    We parameterize each handoff as a package of upstream
    instructions, downstream instructions, and a shared
    communication convention, aligning the optimization unit
    with the dependency between context construction and use.

    \item \textbf{Automated paired refinement.}
    We introduce PACO, combining state-aware edge selection,
    execution-guided two-round revision, and staged end-to-end
    evaluation to optimize these packages under a token budget.

    \item \textbf{Strong performance and cross-dataset transfer.}
    PACO outperforms leading evaluated baselines while keeping
    model weights and the workflow fixed, and retains its
    advantage under transfer without target-specific optimization.
    Ablations and execution analyses provide complementary
    evidence for the effectiveness of paired communication
    optimization.
\end{itemize}
\section{Problem Formulation}
\label{sec:problem}

\subsection{Multi-Agent Systems with Fixed Topology}
\label{sec:mas_setting}

We consider a multi-agent system represented by a fixed directed acyclic graph $G=(V,E)$, where each node $v\in V$ is an LLM-based agent and each edge $(u,v)\in E$ carries task context from agent $u$ to agent $v$. For a task input $x$, let $x_v$ denote the information available to agent
$v$ under its fixed access rules. Agents execute in topological order. The context received by agent $v$ is
\begin{equation}
    C_v = \bigoplus_{u\in\operatorname{Pred}(v)} m_{u\to v},
    \label{eq:incoming_context}
\end{equation}
where $\operatorname{Pred}(v)$ is the set of its immediate predecessors, $m_{u\to v}$ is the message delivered from $u$ to $v$, and $\bigoplus$ denotes concatenation in a fixed order. Messages are obtained from upstream outputs through the workflow's fixed routing rules; source
agents receive no upstream context.

Let $\Theta$ denote the system's communication-instruction configuration. Agent $v$ generates its output according to
\begin{equation}
    o_v \sim f_v\!\left(\,\cdot\mid b_v,p_v(\Theta),x_v,C_v\right),
    \label{eq:agent_execution}
\end{equation}
where $f_v$ is its fixed black-box generation mechanism, $b_v$ contains its fixed role instructions, and $p_v(\Theta)$ denotes the communication instructions induced by $\Theta$. These instructions govern how the agent interprets incoming context and constructs outgoing context. The complete workflow produces a final prediction $\hat y=F_G(x;\Theta,\xi)$, where $\xi$ captures execution randomness.

\subsection{Communication Optimization}
\label{sec:communication_objective}

Let $\Omega$ be the space of admissible communication configurations. Across this space, model weights, agent roles, topology, tool permissions, information access, and output-scoring rules remain fixed. Given a task distribution $\mathcal D$ and a task scorer $q(\hat y,y)\in[0,1]$, our objective is
\begin{equation}
    \begin{aligned}
        \Theta^\star &\in \arg\max_{\Theta\in\Omega} J(\Theta),\\
        J(\Theta) &= \mathbb E_{(x,y)\sim\mathcal D,\,\xi}
        \left[q\!\left(F_G(x;\Theta,\xi),y\right)\right].
    \end{aligned}
    \label{eq:communication_objective}
\end{equation}
Here $y$ denotes reference information available only to the evaluator. The objective measures end-to-end task quality rather than isolated agent outputs or message-format compliance.

In practice, $J$ is estimated through executions on a finite optimization set, separate from the final test set. Search operates under an optimization-token budget $B$ covering candidate generation and workflow evaluation. The budget constrains search expenditure, while task quality
remains the optimization objective. The instructions are discrete, and their effects are assessed through black-box executions. Moreover, the objective is coupled across agents: changing upstream communication alters the context on which downstream reasoning depends. The next section introduces PACO's edge-based parameterization of $\Theta$ and its automated optimization procedure.

\section{Paired Automated Communication Optimization}
\label{sec:method}

We present PACO's communication parameterization and optimization procedure, with the complete algorithm provided in Appendix~\ref{app:algorithm} (Algorithm~\ref{alg:paco}).

\subsection{Paired Communication Parameterization}
\label{sec:interface}

For each edge $e=(u,v)$, PACO defines a communication package
\begin{equation}
    \theta_e=(s_e,r_e,c_e,\ell_e),
    \qquad \Theta=(\theta_e)_{e\in E}.
    \label{eq:interface}
\end{equation}
Its three behavioral components are upstream instructions $s_e$ for constructing task context, downstream instructions $r_e$ for interpreting and using it, and a shared convention $c_e$ specifying common meanings and expectations. The additional attribute $\ell_e$ specifies the message representation, such as natural language or a typed record.

These packages determine the communication instructions $p_v(\Theta)$ introduced in Section~\ref{sec:mas_setting}:
\begin{equation}
    p_v(\Theta)=
    \bigoplus_{e\in\operatorname{In}(v)}\mathsf{Recv}(\theta_e)
    \;\oplus\!
    \bigoplus_{e\in\operatorname{Out}(v)}\mathsf{Send}(\theta_e).
    \label{eq:composition}
\end{equation}
Here $\operatorname{In}(v)$ and $\operatorname{Out}(v)$ are the incoming and outgoing edges of $v$. The fixed renderers $\mathsf{Recv}$ and $\mathsf{Send}$ assemble the corresponding endpoint instructions and shared convention into incoming and outgoing prompt blocks. The blocks
specify how to interpret and construct the chosen representation; representation information may be conveyed through the instructions and convention rather than a standalone $\ell_e$ field. Blocks follow a fixed order, and absent blocks are omitted.

The complete agent input integrates these communication blocks with existing prompt components:
\begin{equation}
    \pi_v^{\mathrm{full}}=
    \mathsf{Assemble}\!\left(
    \pi_v^{\mathrm{fixed}},p_v(\Theta),x_v,C_v\right).
    \label{eq:full_prompt}
\end{equation}
The fixed components include general instructions, the role instructions $b_v$, task-instruction templates, and output requirements. $\mathsf{Assemble}$ denotes the workflow's fixed prompt-construction procedure. The templates remain unchanged during optimization, while
the task input $x_v$ and upstream context $C_v$ vary across executions. Communication modules enter the agents' existing model calls.

An ordinary update to $e=(u,v)$ permits coordinated changes to $u$'s outgoing guidance, $v$'s incoming guidance, and their shared convention, while keeping other instruction blocks fixed. Agent outputs may still change as context propagates. Each edge package is proposed, installed, and evaluated as a unit; individual fields need not change in every candidate. The conventions
are soft: intermediate format diagnostics do not suppress otherwise available message text.

\subsection{Candidate Pool and Selection}
\label{sec:pool}

Inspired by GEPA's candidate-pool design \citep{gepa2026}, PACO maintains a pool $\mathcal{P}$ of complete communication configurations, each comprising the packages of all edges. The pool starts with the initial configuration $\Theta_0$ and retains evaluated configurations with complementary performance on the internal validation set $H$ drawn from the optimization data. This preserves multiple starting points for refinement, including configurations that excel on different examples.

Among configurations satisfying the pool's quality and uniqueness criteria, each example in $H$ credits those attaining its best score. Redundant configurations are pruned while preserving coverage, and PACO samples a configuration $\Theta\in\mathcal{P}$ in proportion to its credits. The selected configuration serves as the parent for the next refinement step. The evaluation procedure is described in Section~\ref{sec:evaluation}, with detailed pool admission and pruning rules provided in Appendix~\ref{app:algorithm}.

\subsection{State-Aware Edge Scheduling}
\label{sec:scheduling}

Given a configuration $\Theta$ selected from the candidate pool, PACO treats each communication edge as a bandit arm and selects the edge to optimize using a rule inspired by discounted UCB \citep{garivier2011switching}. The scheduler balances revisiting edges whose revisions improved task performance with exploring edges that have been assessed less extensively.

For each configuration--edge pair $(\Theta,e)$, the scheduler records the end-to-end gains $\delta_j$ from previous refinement attempts, each estimated from $n_j$ examples. Older observations receive less weight through $w_j=\lambda^{d_j}$, where $0<\lambda<1$ and $d_j$ counts the ordinary search steps taken from $\Theta$ since observation $j$. The discounted reward sum and observation count are
\begin{equation}
    S_{\Theta,e}=\sum_j w_j\delta_j,
    \qquad
    N_{\Theta,e}=\sum_j w_j.
\end{equation}
To account for differences in evaluation batch size, we also define $Q_{\Theta,e}=\sum_j w_j g(n_j)$, where $g(n_j)$ assigns greater exploration weight to observations based on fewer examples. PACO selects the edge with the highest priority:
\begin{equation}
    \begin{aligned}
    e^\star &\in \arg\max_{e\in E} I_{\Theta,e},\\
    I_{\Theta,e} &=
    \underbrace{
        \frac{S_{\Theta,e}}{N_{\Theta,e}+\rho}
    }_{\text{estimated gain}}
    +\alpha
    \underbrace{
        \sqrt{
        \frac{\log\!\left(2+\sum_{a\in E}N_{\Theta,a}\right)
        (Q_{\Theta,e}+\rho)}
        {(N_{\Theta,e}+\rho)^2}}
    }_{\text{exploration bonus}}.
    \end{aligned}
    \label{eq:edge_index}
\end{equation}
Here $\rho>0$ regularizes sparse observations and $\alpha>0$ controls exploration. The first term favors edges with higher discounted gains; the second encourages exploration based on the amount and evaluation batch sizes of past observations. The batch-size adjustment is heuristic rather than a calibrated confidence bound.

Histories are maintained separately for each configuration because an edge's gains depend on the surrounding communication instructions. New configurations start with zero statistics, and discounting advances only when that configuration is selected for an ordinary search step. Rewards come from the staged evaluations in Section~\ref{sec:evaluation}. The specific discount factor, batch-size correction, and reward rules are given in Appendix~\ref{app:bandit}.

\subsection{Two-Round Refinement}
\label{sec:revision}

After selecting an edge $e$, PACO refines its communication package in two rounds. Feedback from the parent alone cannot reveal how proposed instructions will interact during execution. The second round therefore uses first-round execution feedback to address remaining or newly introduced mismatches in context construction and use. Each round may revise any subset of the package's endpoint instructions, shared convention, and message representation, while proposing and evaluating the package as a unit. For an ordinary step, edits are restricted to $e$, and the parent $\Theta$ remains the evaluation reference throughout.

The parent first executes on a small pilot batch $D_{\mathrm{p}}$. An LLM-based proposer uses the current package, fixed runtime constraints, and execution feedback to generate $K_1$ candidate configurations $\{\Theta^{(1,k)}\}_{k=1}^{K_1}$, differing from the parent only within $e$. Valid candidates execute on the same pilot batch. The best scorable first-round candidate becomes $\Theta_{\mathrm{base}}$. The proposer then generates $K_2$ revisions $\{\Theta^{(2,k)}\}_{k=1}^{K_2}$ from this base, informed by executions of the parent and first-round candidates. Valid candidates from both rounds remain eligible for subsequent selection.

The generation base and evaluation reference serve different purposes. Even if every first-round candidate regresses, the best scorable one can seed further refinement; all candidates are still assessed against the original parent. Invalid, duplicate, or unchanged candidates are
filtered out, so the evaluated set may contain fewer than $K_1+K_2$ candidates. If no first-round candidate is scorable, the second round is skipped.

Feedback includes agent-visible inputs, exchanged messages, outcome categories, and eligible earlier execution cases. Reference answers, hidden reasoning, and per-example numerical verifier scores are excluded from proposer feedback. Within each refinement step, promotion evaluations occur after proposal generation and guide candidate selection. Their executions may inform later steps through historical search-set feedback, whereas evaluations on $H$ are never used as proposal examples. We use $K_1=K_2=2$ in our experiments.

\subsection{Staged End-to-End Evaluation}
\label{sec:evaluation}

Candidates from both refinement rounds are evaluated within the complete workflow, capturing the downstream effects of communication revisions. Let $q_i(\Theta)$ be the observed task score on example $i$. The estimated gain of candidate $\Theta'$ relative to the original parent $\Theta$ on an evaluation batch $D$ is
\begin{equation}
    \Delta_D(\Theta';\Theta)=
    \frac{1}{|D|}\sum_{i\in D}
    \left[q_i(\Theta')-q_i(\Theta)\right].
    \label{eq:gain}
\end{equation}
Scores are paired by task example, while communication packages outside the permitted edit scope remain unchanged from the parent.

Evaluation proceeds through three stages. Disjoint pilot and promotion batches are drawn from search data separate from the internal validation set $H$. The pilot batch $D_{\mathrm{p}}$ screens candidates from both rounds against a positive-gain threshold. A few survivors advance to the promotion batch $D_{\mathrm{m}}$, where gains use only the new examples. The candidate with the highest positive promotion gain is then evaluated on $H$. This concentrates evaluation effort on promising candidates, guided throughout by end-to-end task performance.

Configurations with complete evaluations on $H$ are archived and used to refresh the candidate pool (Section~\ref{sec:pool}). Near-best configurations with complementary coverage may be retained even when their mean score is lower than their parent's. Since $H$ guides search and configuration selection, it belongs to the optimization data and remains separate from the final test set.

Each ordinary refinement step supplies at most one reward to the selected configuration's edge bandit (Section~\ref{sec:scheduling}), computed from the deepest valid evaluation stage. Evaluation settings and implementation details are provided in Appendix~\ref{app:algorithm}.

\subsection{Cooperative Refinement}
\label{sec:cooperative}

Improving one edge may enable further coordination across the workflow. When an ordinary single-edge step achieves a new best mean score on the full internal validation set $H$, PACO triggers one cooperative refinement step, using the improved configuration as its parent. This step may revise multiple edge packages, following the same two-round refinement, staged evaluation, and pool admission rules. It neither updates the single-edge bandit nor triggers further cooperative refinement.

The overall search operates under a shared optimization-token budget. At its stopping boundary, PACO completes ongoing steps and freezes the configuration with the highest mean score on the full validation set $H$ for testing. Algorithm~\ref{alg:paco} in Appendix~\ref{app:algorithm}
summarizes the complete procedure.

\section{Experiments}
\label{sec:experiments}

\subsection{Experimental Setup}
\label{sec:setting}

\paragraph{Tasks and data.}
We evaluate on HotpotQA \citep{yang-etal-2018-hotpotqa} and MuSiQue-Ans \citep{trivedi2022musique} for multi-hop reasoning, and DROP \citep{dua2019drop} for discrete reasoning over text. Optimization uses 500 HotpotQA examples: 400 for sampling pilot and promotion batches ($D_{\mathrm{p}},D_{\mathrm{m}}$) and 100 for internal validation ($H$). After optimization, the best configuration on $H$ is frozen for evaluation on 800 HotpotQA questions and 400 questions per transfer dataset, without further optimization. Appendix~\ref{app:data} details data construction.

\paragraph{Workflow and initialization.}
We use a fixed four-agent chain: Scout $\rightarrow$ Resolver $\rightarrow$ Integrator $\rightarrow$ Verifier. All agents receive the question, but only the first two receive separate evidence shards. Messages are accessible only to the direct successor, making downstream reasoning dependent on context preserved across handoffs. PACO starts from the unoptimized (vanilla) workflow, with simple, generic instructions for context construction and use and basic
edge-specific conventions suggesting typed records with soft formatting requirements. Role instructions and information access remain fixed throughout optimization. Appendix~\ref{app:workflow} provides the complete configuration.

\paragraph{Models and optimization.}
Workflow agents and the LLM-based proposer use \texttt{gpt-5.6-luna} with low and high reasoning effort, respectively. PACO uses a 25M-token optimization budget and optimizer seed 0, with model weights fixed throughout. Search hyperparameters and the complete algorithm appear in Appendix~\ref{app:algorithm}.

\paragraph{Baselines and metrics.}
We compare with the unoptimized (vanilla) workflow; message-processing and communication approaches, including Faithful Rewriter and Task-Aware Rewriter with general preservation
and task-specific instructions, respectively, LLMLingua-2 \citep{pan2024llmlingua2}, OPTiMACS \citep{gupta2026optimacs}, and PACT \citep{huang2026pact}; and prompt optimizers MASPO, MASPOB, and GEPA \citep{wang2026maspo,hong2026maspob,gepa2026}. We report Joint, Answer, and Supporting Facts F1/EM on HotpotQA, Answer F1/EM for transfer, and optimization/deployment tokens in millions. Failed or invalid predictions receive zero, with all questions retained in the denominator. Implementation and accounting details appear in Appendices~\ref{app:baselines}.

\begin{table}[t]
  \centering
  \caption{\textbf{HotpotQA performance and cost.} Scores are percentages; parentheses show deployment-token changes relative to vanilla. Methods are grouped by optimization approach. Support denotes supporting-fact evaluation. Bold and underlined values mark the best and second-best scores.}
  \label{tab:hotpot}
  \begingroup
  \fontsize{8.5}{10}\selectfont
  \setlength{\tabcolsep}{2pt}
  \begin{tabular*}{\linewidth}
    {@{\extracolsep{\fill}}lrrrrrrrr@{}}
    \toprule
    & \multicolumn{2}{c}{Joint $\uparrow$}
    & \multicolumn{2}{c}{Answer $\uparrow$}
    & \multicolumn{2}{c}{Support $\uparrow$}
    & \multicolumn{2}{c}{Tokens (M)} \\
    \cmidrule(lr){2-3}
    \cmidrule(lr){4-5}
    \cmidrule(lr){6-7}
    \cmidrule(l){8-9}
    Method & F1 & EM & F1 & EM & F1 & EM & Opt. & Deploy. \\
    \midrule
    Vanilla
      & 30.00 & 8.88 & 41.64 & 29.63
      & 38.68 & 15.75 & 0.00 & 3.91 \\
    \midrule
    Faithful Rewriter
      & 29.36 & 8.25 & 41.16 & 30.63
      & 37.95 & 14.75 & 0.00 & 6.84 (+75\%) \\
    Task-Aware Rewriter
      & 24.35 & 7.00 & 33.82 & 25.00
      & 30.92 & 11.38 & 0.00 & 8.60 (+120\%) \\
    LLMLingua-2 \citep{pan2024llmlingua2}
      & 29.36 & 8.50 & 41.15 & 29.50
      & 37.74 & 13.88 & 0.46 & 3.90 ($-0.2$\%) \\
    OPTiMACS \citep{gupta2026optimacs}
      & 24.37 & 6.88 & 34.18 & 24.50
      & 31.35 & 12.00 & 19.10 & 8.63 (+121\%) \\
    PACT \citep{huang2026pact}
      & 46.43 & 11.63 & 65.51 & 47.38
      & 58.46 & 20.50 & 0.00 & 4.10 (+5\%) \\
    \midrule
    MASPO \citep{wang2026maspo}
      & 37.97 & 17.13 & 46.35 & 36.50
      & 45.78 & 25.50 & 24.27 & 7.23 (+85\%) \\
    MASPOB \citep{hong2026maspob}
      & 47.35 & 17.63 & 59.00 & 42.00
      & 61.48 & 31.50 & 24.74 & 4.12 (+5\%) \\
    GEPA \citep{gepa2026}
      & \underline{59.27} & \underline{27.00}
      & \underline{68.80} & \underline{50.00}
      & \underline{74.91} & \textbf{46.13}
      & 24.20 & 5.25 (+34\%) \\
    \midrule
    \textbf{PACO (Ours)}
      & \textbf{61.26} & \textbf{28.88}
      & \textbf{72.15} & \textbf{54.75}
      & \textbf{75.75} & \underline{45.88}
      & 23.96 & 5.21 (+33\%) \\
    \bottomrule
  \end{tabular*}
  \endgroup
\end{table}

\begin{table}[t]
  \centering
  \caption{\textbf{Transfer without further optimization.} HotpotQA configurations are frozen for transfer. F1/EM measure answer quality; units and highlighting follow Table~\ref{tab:hotpot}.}
  \label{tab:transfer}
  \begingroup
  \fontsize{8.5}{10}\selectfont
  \setlength{\tabcolsep}{2pt}
  \begin{tabular*}{\linewidth}
    {@{\extracolsep{\fill}}lrrrrrr@{}}
    \toprule
    & \multicolumn{3}{c}{DROP}
    & \multicolumn{3}{c}{MuSiQue-Ans} \\
    \cmidrule(lr){2-4}
    \cmidrule(l){5-7}
    Method & F1 $\uparrow$ & EM $\uparrow$ & Deploy.
           & F1 $\uparrow$ & EM $\uparrow$ & Deploy. \\
    \midrule
    Vanilla
      & 42.25 & 21.50 & 1.28
      & 29.60 & 21.25 & 2.43 \\
    \midrule
    PACT \citep{huang2026pact}
      & 45.46 & \underline{27.25} & 1.47 (+15\%)
      & 37.94 & 28.50 & 2.52 (+4\%) \\
    \midrule
    MASPO \citep{wang2026maspo}
      & 15.70 & 9.50 & 2.84 (+122\%)
      & 25.12 & 19.50 & 4.08 (+68\%) \\
    MASPOB \citep{hong2026maspob}
      & 42.80 & 20.50 & 1.38 (+8\%)
      & 32.83 & 22.75 & 2.54 (+5\%) \\
    GEPA \citep{gepa2026}
      & \underline{49.09} & 21.00 & 1.92 (+50\%)
      & \underline{44.50} & \underline{31.00} & 3.12 (+28\%) \\
    \midrule
    \textbf{PACO (Ours)}
      & \textbf{54.95} & \textbf{35.75} & 1.86 (+45\%)
      & \textbf{45.06} & \textbf{35.75} & 3.11 (+28\%) \\
    \bottomrule
  \end{tabular*}
  \endgroup
\end{table}

\subsection{In-Domain Results}
\label{sec:hotpot}

Table~\ref{tab:hotpot} shows that PACO leads on five of six quality metrics, gaining 31.26 Joint F1 points over vanilla and 1.98 over GEPA, the strongest baseline. Gains span answer quality and Supporting Facts F1, while GEPA retains a slight advantage in Supporting Facts EM. Prompt optimizers and PACT substantially outperform vanilla, whereas rewriting and compression do not improve Joint F1. Modifying messages without adapting downstream instructions may discard useful context or hinder its use. Among prompt optimizers, GEPA outperforms MASPOB and MASPO, potentially reflecting differences in candidate generation and evaluation. MASPO's LLM-judged rewards may diverge from the task metric; MASPOB uses task scores but searches predefined prompt sets. GEPA combines task-score-based evaluation with execution-informed prompt evolution and diverse candidate selection. PACO achieves higher task performance than GEPA while using slightly fewer tokens for both optimization and deployment.

\subsection{Cross-Dataset Transfer}
\label{sec:transfer}

Table~\ref{tab:transfer} assesses the cross-dataset generalization of communication skills learned on HotpotQA, with configurations frozen and no target-specific optimization. PACO leads in Answer F1 and EM on both datasets, exceeding GEPA by 5.86 F1 points on DROP and 0.56 on MuSiQue-Ans. Its advantage on DROP suggests that these skills extend beyond multi-hop reasoning to discrete reasoning over text. In contrast, MASPO falls below vanilla on both datasets despite improving in-domain, showing that source-task gains do not necessarily transfer. PACO retains its advantage with slightly fewer deployment tokens than GEPA on both datasets, supporting the generalization of its communication skills without increased deployment cost.

\subsection{Ablation Study}
\label{sec:ablation}

All ablations use a 5M-token optimization budget, with other experimental settings unchanged.
Table~\ref{tab:ablation-pairing} examines the scope of communication refinement. Upstream-only and downstream-only variants fix the opposite endpoint and shared convention, while independent updates optimize all three components separately. PACO outperforms the two single-endpoint variants by 2.06 and 3.11 Joint F1 points, respectively. The stronger upstream-only result is consistent with the importance of supplying useful context, but improving either endpoint alone leaves gains unrealized. Independent updates narrow the gap yet remain 0.93 Joint F1 points below PACO despite using more optimization tokens, supporting the value of coordinated revisions.

Table~\ref{tab:ablation-search} examines edge scheduling and two-round refinement. Uniform edge selection averages 41.41 Joint F1 across three runs, compared with PACO's seed-0 result of 49.35 at similar optimization cost. Uniform selection occasionally yields much lower scores, whereas repeated PACO runs span less than 1.5 Joint F1 points. These observations suggest that adaptive scheduling makes optimization more reliable under a limited budget. Single-round refinement comes closer to PACO but scores 0.78 Joint F1 points lower, with a smaller gap in Answer F1, consistent with a modest benefit from first-round execution feedback.

\begin{table}[t]
  \centering
  \begin{minipage}[t]{0.49\linewidth}
    \centering
    \caption{\textbf{Effect of paired refinement.} Opt. denotes optimization tokens in millions.}
    \label{tab:ablation-pairing}
    \begingroup
    \fontsize{8.5}{10}\selectfont
    \setlength{\tabcolsep}{1.5pt}
    \begin{tabular*}{\linewidth}
      {@{\extracolsep{\fill}}lccc@{}}
      \toprule
      Variant & Joint F1 & Answer F1 & Opt. (M) \\
      \midrule
      PACO & 49.35 & 59.36 & 4.57 \\
      \midrule
      Upstream only & 47.29 & 56.78 & 4.78 \\
      Downstream only & 46.24 & 55.71 & 4.67 \\
      Independent updates & 48.42 & 58.25 & 4.91 \\
      \bottomrule
    \end{tabular*}
    \endgroup
  \end{minipage}
  \hfill
  \begin{minipage}[t]{0.49\linewidth}
    \centering
    \caption{\textbf{Effect of search components.} Notation follows Table~\ref{tab:ablation-pairing}. $^{*}$ denotes means over three runs.}
    \label{tab:ablation-search}
    \begingroup
    \fontsize{8.5}{10}\selectfont
    \setlength{\tabcolsep}{1.5pt}
    \begin{tabular*}{\linewidth}
      {@{\extracolsep{\fill}}lccc@{}}
      \toprule
      Variant & Joint F1 & Answer F1 & Opt. (M) \\
      \midrule
      PACO & 49.35 & 59.36 & 4.57 \\
      \midrule
      Uniform edge selection & ~~41.41$^{*}$ & ~~51.02$^{*}$ & ~~4.63$^{*}$ \\
      Single-round refinement & 48.57 & 59.11 & 4.55 \\
      \bottomrule
    \end{tabular*}
    \endgroup
  \end{minipage}
\end{table}

\subsection{Case Study}
\label{sec:case-study}

Table~\ref{tab:case-study} compares vanilla and frozen PACO on the same question, evidence shards, access permissions, and evaluation seed. Vanilla transmits the complete trade sentence, yet the Resolver reverses the transaction direction and the error persists downstream. PACO makes both destinations explicit and combines the handoff with local award evidence to answer correctly. Its learned sender instructions require each fact's ``subject, relation, object,'' while the receiver checks each link using incoming and local evidence. Both runs return identical citations, so the score difference comes from answer correctness. This illustrates aligned context construction and use, without attributing the improvement to any single package component (Appendix~\ref{app:case-study}).

\begin{table}[t]
  \centering
  \caption{\textbf{Making transaction direction explicit.} Selected message excerpts; surrounding text is omitted. Both final predictions are valid. F1 scores are percentages.}
  \label{tab:case-study}
  \begingroup
  \fontsize{8.5}{10}\selectfont
  \setlength{\tabcolsep}{4pt}
  \renewcommand{\arraystretch}{1.08}
  \begin{tabular}{@{}>{\raggedright\arraybackslash}p{0.16\linewidth}>{\raggedright\arraybackslash}p{0.395\linewidth}>{\raggedright\arraybackslash}p{0.395\linewidth}@{}}
    \toprule
    \multicolumn{3}{@{}p{\linewidth}@{}}{\textbf{Question:}
    To what team was the 2014 NBA Rookie of the Year traded in October 2016?} \\
    \multicolumn{3}{@{}p{\linewidth}@{}}{\textbf{Resolver's local evidence:}
    Michael Carter-Williams was named NBA Rookie of the Year in 2014.} \\
    \midrule
    & \textbf{Vanilla} & \textbf{PACO} \\
    \midrule
    Scout handoff
    & ``The Chicago Bulls traded Tony Snell to the Milwaukee Bucks for
      Michael Carter-Williams on October 17, 2016.''
    & ``Tony Snell $\rightarrow$ Milwaukee Bucks; Michael Carter-Williams
      $\rightarrow$ Chicago Bulls'' \\
    Resolver output
    & ``the incoming claim states he was traded to the Milwaukee Bucks
      in October 2016.''
    & ``therefore, Carter-Williams was traded to Chicago'' \\
    Final answer
    & Milwaukee Bucks & \textbf{Chicago Bulls} \\
    Answer / Joint F1
    & 0 / 0 & \textbf{100 / 80} \\
    \bottomrule
    \multicolumn{3}{@{}p{\linewidth}@{}}{\footnotesize
    Both score 80 Supporting Facts F1, citing the same two facts and missing the third.} \\
  \end{tabular}
  \endgroup
\end{table}

\section{Related Work}
\label{sec:related-work}

\paragraph{Workflow and topology optimization.}
Automated MAS design optimizes both agent behavior and system structure.
GPTSwarm refines node prompts and graph connectivity
\citep{zhuge2024gptswarm}, while AFlow and MaAS automate workflow
construction and query-dependent architecture selection, respectively
\citep{aflow2025,maas2025}. MASS interleaves prompt and topology
optimization \citep{zhou2026mass}, and AgentPrune removes redundant
connections from communication graphs \citep{zhang2025agentprune}.
PACO complements these approaches by optimizing the communication
instructions on existing edges while keeping topology fixed.

\paragraph{Prompt optimization for compound systems.}
DSPy and MIPRO optimize modular LM programs through demonstrations and
instructions \citep{khattab2024dspy,opsahlong2024mipro}; TextGrad and
GEPA use textual feedback and execution reflection to refine system
components, respectively \citep{yuksekgonul2025textgrad,gepa2026}.
For MAS, MASPO uses topology-aware prompt evolution with multilevel
feedback, while MASPOB searches prompt combinations using a graph-based
bandit surrogate \citep{wang2026maspo,hong2026maspob}.
MAPRO explicitly models agent quality and handoff compatibility,
combining MAP inference with topology-aware feedback to coordinate
agent-prompt selection and refinement \citep{zhang2026mapro}.
These methods account for component dependencies, with joint changes
supported by GEPA's merge operation and MASPOB's block variant.
PACO's distinction is its optimization unit: the
upstream instructions, downstream instructions, and shared convention
of a communication edge are jointly proposed and evaluated as one package.

\paragraph{Communication protocols and learning.}
Communication can also be improved through message design, compression,
and model training. PACT structures messages as compact action-state
records \citep{huang2026pact}, while OPTiMACS learns to select task-dependent
message representations \citep{gupta2026optimacs}. LLMLingua-2 learns
token-level prompt compression \citep{pan2024llmlingua2}, and Optima
trains agents to balance task performance, communication efficiency,
and readability \citep{chen2025optima}.
PACO learns communication skills through the instructions of black-box
agents, requiring neither model updates nor additional message-rewriting
calls.
\section{Limitations}
\label{sec:limitations}

\textbf{Evaluation scope.}
Experiments cover three question-answering benchmarks with one model family and a fixed four-agent chain; broader workflows remain untested.
\textbf{Feedback requirements.}
PACO relies on reliable task feedback and repeated end-to-end evaluations, limiting its use when evaluation is difficult or costly.
\textbf{Deployment cost.}
PACO prioritizes quality and increases deployment tokens over vanilla. Jointly optimizing quality, cost, and latency remains future work.

\section{Conclusion}
\label{sec:conclusion}

We presented PACO, an edge-centric framework that jointly optimizes upstream instructions, downstream instructions, and their shared communication convention in black-box multi-agent systems. With model weights and topology fixed, PACO combines adaptive edge selection with execution-guided paired refinement and end-to-end candidate evaluation. Experiments demonstrate substantial improvements over vanilla and consistent gains over leading baselines, including GEPA. The learned communication skills retain their advantage across transfer benchmarks without target-specific optimization. Ablations support coordinated updates and adaptive edge selection under limited budgets, while the case study illustrates improved expression and interpretation of task-relevant relations. Together, these findings establish edge-centric paired optimization as an effective approach to multi-agent collaboration, aligning what agents convey with how their successors interpret and use it.
\clearpage
\subsection*{AI Use Statement}
We used AI assistance to support manuscript preparation, including drafting, language polishing, LaTeX editing. The authors are responsible for verifying the accuracy of the text, references, and reported findings, and take full responsibility for the final manuscript.

\subsection*{Ethics Statement}
This study uses existing question-answering benchmarks and does not collect new data from human participants. PACO optimizes communication instructions while preserving agents' predefined information access and tool permissions. Improved benchmark performance does not guarantee factual correctness or safe autonomous behavior, as erroneous information may still propagate through agent interactions.

\subsection*{Reproducibility Statement}
Sections~\ref{sec:method} and~\ref{sec:setting} describe the method and experimental setup, with additional implementation and evaluation details in the appendix. We specify the workflow, model settings, data splits, optimization budgets, and scoring procedures. We will release the source code upon acceptance. Because execution relies on stochastic black-box models, identical configurations may not produce identical outputs across repeated runs.

\bibliography{references}
\bibliographystyle{iclr2027_conference}

\clearpage
\appendix
\section{Optimization Procedure and Hyperparameters}
\label{app:algorithm}

Algorithm~\ref{alg:paco} summarizes the procedure in
Section~\ref{sec:method}. The archive $\mathcal A$ contains
configurations with complete evaluations on the 100-example
internal validation set $H$. The active pool
$\mathcal P\subseteq\mathcal A$ supplies parents for refinement.
Both search data and $H$ belong to the optimization data;
final test results are not used for configuration selection.

\begin{algorithm}[h]
\caption{Paired Automated Communication Optimization (PACO)}
\label{alg:paco}
\small
\begin{algorithmic}[1]
\Require Fixed workflow $G$, initial configuration $\Theta_0$, task metric $q$
\Require Search data $D_{\mathrm{search}}$, internal validation set $H$, token budget $B$
\Require Candidate counts $K_1,K_2$
\Ensure Frozen communication configuration $\widehat\Theta$

\State Evaluate $\Theta_0$ on $H$;
       $\mathcal{A}\gets\{\Theta_0\}$;
       $\mathcal{P}\gets\{\Theta_0\}$
\State Initialize configuration-specific bandit statistics;
       $T\gets\varnothing$

\While{the search budget admits another step}
    \If{$T\neq\varnothing$}
        \State $\Theta\gets\Call{ConsumeTrigger}{T}$;
               $T\gets\varnothing$
        \State $E_{\mathrm{edit}}\gets E$;
               $c\gets\mathrm{true}$
               \Comment{Cooperative step}
    \Else
        \State $\Theta\gets\Call{SelectCandidate}{\mathcal{P}}$
        \State $e\gets\arg\max_{a\in E} I_{\Theta,a}$
        \State $E_{\mathrm{edit}}\gets\{e\}$;
               $c\gets\mathrm{false}$
               \Comment{Ordinary step}
    \EndIf

    \State $(D_{\mathrm{p}},D_{\mathrm{m}})
           \gets\Call{PrepareBatches}{D_{\mathrm{search}},\Theta,c}$
    \State Obtain or reuse compatible parent pilot executions
           and feedback $F_\Theta$
    \State $(\mathcal{C}_1,F_1)
           \gets\Call{PilotCandidates}
           {\Theta,E_{\mathrm{edit}},K_1,D_{\mathrm{p}},F_\Theta}$
    \State $\mathcal{C}_2\gets\varnothing$
    \If{$\mathcal{C}_1\neq\varnothing$}
        \Comment{Scorable candidates, including nonpositive gains}
        \State $\Theta_{\mathrm{base}}
               \gets\arg\max_{\Psi\in\mathcal{C}_1}
               \bar q_{D_{\mathrm{p}}}(\Psi)$
        \State $(\mathcal{C}_2,F_2)
               \gets\Call{PilotCandidates}
               {\Theta_{\mathrm{base}},E_{\mathrm{edit}},K_2,
                D_{\mathrm{p}},(F_\Theta,F_1)}$
    \EndIf

    \State $b\gets\max_{\Psi\in\mathcal{A}}\bar q_H(\Psi)$
    \State $\mathcal{C}\gets\Call{PilotScreen}
           {\mathcal{C}_1\cup\mathcal{C}_2,\Theta,D_{\mathrm{p}}}$
    \State $\Theta'\gets\Call{PromotionSelect}
           {\mathcal{C},\Theta,D_{\mathrm{m}}}$
    \If{$\Theta'\neq\varnothing$}
        \State Evaluate $\Theta'$ on $H$
    \EndIf
    \State Retain all stage evaluation records,
           including those of screened-out candidates

    \If{$\Theta'\neq\varnothing$
        \textbf{and} its evaluation on $H$ is complete}
        \State $\mathcal{A}\gets\mathcal{A}\cup\{\Theta'\}$;
               $\mathcal{P}\gets\Call{RefreshPool}{\mathcal{A}}$
        \State Initialize zero bandit statistics for any new configuration
        \If{$\neg c$ \textbf{and} $\bar q_H(\Theta')>b$}
            \State $T\gets\Call{CreateTrigger}{\Theta'}$
        \EndIf
    \EndIf

    \If{$\neg c$}
        \State Discount $\Theta$'s edge statistics;
               add at most one reward to edge $e$
        \Statex \hspace{\algorithmicindent}\hspace{\algorithmicindent}
                using retained records from the deepest valid evaluation stage
    \EndIf
\EndWhile

\State $\widehat\Theta\gets
       \arg\max_{\Psi\in\mathcal{A}}\bar q_H(\Psi)$
\State Freeze $\widehat\Theta$ for testing
\State \Return $\widehat\Theta$
\end{algorithmic}
\end{algorithm}

\noindent
\textsc{PilotCandidates} returns admissible candidates with
available pilot scores, including candidates with nonpositive
gains. The second-round base is used only for proposal generation;
all gains are measured against the original parent $\Theta$.
Screening does not discard evaluation records needed for the
bandit update. If no valid gain is available, the ordinary step
applies discounting only.

\begin{table}[t]
\centering\small
\caption{\textbf{PACO optimization settings.}
Thresholds use scores in $[0,1]$. Settings describe the main
25M-token run; ablations use a 5M-token budget.}
\label{tab:optimization-settings}
\begin{tabular*}{\linewidth}{@{\extracolsep{\fill}}lr@{}}
\toprule
Setting & Value \\
\midrule
Workflow / proposer model
  & \texttt{gpt-5.6-luna} / \texttt{gpt-5.6-luna} \\
Workflow / proposer reasoning effort & Low / high \\
Proposer maximum output tokens & 16,384 \\
Optimizer seed & 0 \\
Search / internal validation examples & 400 / 100 \\
Pilot / promotion examples per step & 4 / 8 \\
Candidates per refinement round, $K_1,K_2$ & 2, 2 \\
Maximum promotion candidates per step & 2 \\
Maximum candidates entering $H$ per step & 1 \\
Pilot gain threshold & $\Delta_{D_{\mathrm p}}>0.01$ \\
Promotion gain threshold & $\Delta_{D_{\mathrm m}}>0$ \\
Candidate-pool quality margin & 0.05 \\
Maximum active-pool size & Unbounded \\
Early-rejection margin on $H$ & 0.05 \\
Evaluation batch size on $H$ & 32 \\
Maximum historical execution examples & 2 \\
Maximum historical memory entries & 2 \\
Nominal optimization budget & 25,000,000 tokens \\
New-step stopping threshold & 23,750,000 tokens \\
\bottomrule
\end{tabular*}
\end{table}

\paragraph{Candidate-pool construction.}
Let
$q^\star=\max_{\Psi\in\mathcal A}\bar q_H(\Psi)$
be the best archived mean Joint F1. The quality-eligible set is
\begin{equation}
    \mathcal A_{\mathrm{qual}}
    =
    \left\{
    \Theta\in\mathcal A:
    \bar q_H(\Theta)\geq q^\star-0.05
    \right\}.
\end{equation}
The boundary is inclusive. Scores are compared using exact
rational representations of the recorded numerical values,
without a floating-point tolerance or display rounding.

For each example $i\in H$, define
\begin{equation}
    M_i=\max_{\Psi\in\mathcal A_{\mathrm{qual}}}q_i(\Psi),
    \qquad
    W_\Theta=\{i\in H:q_i(\Theta)=M_i\}.
\end{equation}
Every configuration tied for the best score receives one full
win for that example; credit is not divided among tied
configurations.

To remove redundant configurations, initialize
$C_i=|\{\Theta\in\mathcal A_{\mathrm{qual}}:i\in W_\Theta\}|$.
Visit configurations in ascending order of
$(\bar q_H(\Theta),\operatorname{hash}(\Theta))$.
Remove a configuration if $C_i>1$ for every $i\in W_\Theta$,
decrementing those coverage counts after removal.
The resulting active pool preserves at least one winner
for every example and has no explicit size limit.
Parent sampling uses
\begin{equation}
    \Pr(\Theta\mid\mathcal P)
    =
    \frac{|W_\Theta|}
         {\sum_{\Psi\in\mathcal P}|W_\Psi|}.
\end{equation}
Because ties receive full credit, the denominator can exceed
$|H|$. Archived configurations and their bandit statistics
are retained when they leave the active pool.

\paragraph{Batch construction and paired refinement.}
An ordinary step samples 12 distinct examples from the
400-example search set using a step-specific random-number
domain: four form $D_{\mathrm p}$ and eight form
$D_{\mathrm m}$. The original parent is the evaluation
reference throughout both refinement rounds.

Each proposer call requests two candidates. Invalid,
duplicate, and unchanged configurations are discarded
without rollout or replacement. Admissible candidates
execute on the pilot batch. If at least one first-round
candidate is scorable, the best candidate seeds the second
round, even when all first-round gains are nonpositive;
otherwise, the second round is skipped.
Candidates from both rounds remain eligible for screening.
The generation base changes in the second round, but
candidate gains are always measured against the original
parent rather than accumulated across revisions.

\paragraph{Proposal feedback.}
The proposer receives task questions, agent-visible evidence
and inputs, exchanged messages, recorded agent outputs,
and outcome or communication diagnostics.
Reference answers, hidden reasoning, and per-example
numerical task scores are excluded. Paired score differences
are represented only by the directions
\texttt{improved}, \texttt{regressed}, or \texttt{unchanged};
individual examples may include a task-failure indicator.
Recorded model answers remain part of the visible executions.

Current parent-pilot executions provide the main feedback,
and the second round additionally uses first-round candidate
executions. Up to two historical execution examples may be
drawn from completed pilot or promotion evaluations in earlier
steps of the same run. These examples must belong to the
400-example search set and the current protocol; engineering
runs and cross-run histories are excluded. Selection prioritizes
one improved and one regressed paired example, filling unused
slots with nonperfect individual executions.
Up to two earlier-step memory summaries may also be included,
ranked by configuration and edge applicability and recency.
These summaries do not contain task content.

Promotion executes after proposal generation and therefore
does not provide examples for the current step, but its
records may inform later steps. Evaluations on $H$ are never
indexed as proposal examples. Historical retrieval has no
fixed age limit, and there is no separate feedback-token cap.

\paragraph{Staged evaluation and ranking.}
Candidates are ranked by paired gain in descending order,
with configuration hash in ascending order breaking ties.
This rule applies to the first-round generation base,
pilot screening, and promotion selection.
At most two candidates with
$\Delta_{D_{\mathrm p}}>0.01$
advance to promotion. The highest-ranked candidate with
$\Delta_{D_{\mathrm m}}>0$
proceeds to $H$. Both gates are strict, and promotion gains
use only the eight new promotion examples.

Evaluation records are retained even for screened-out
candidates. An ordinary step contributes at most one bandit
reward, using the highest-ranked candidate at the deepest
stage with a valid paired gain: $H$, then promotion, then
pilot. A partial $H$ evaluation uses the observed prefix
and the parent's scores on the same examples.
Shallower-stage candidates do not replace an available
deeper-stage reward. If no valid gain is available, the
step applies discounting without adding a reward.
Appendix~\ref{app:bandit} specifies the scheduler update.

\paragraph{Early rejection on internal validation.}
For a candidate entering $H$, let $b_{\mathrm{pool}}$ be
the best complete-$H$ mean in the parent-pool snapshot
associated with that step, and fix
$L=b_{\mathrm{pool}}-0.05$.
After observing $m$ examples with total score $T_m$,
the largest possible full-set mean is
\begin{equation}
    U_m=\frac{T_m+|H|-m}{|H|}.
    \label{eq:upper}
\end{equation}
Evaluation proceeds in batches of eight, with a final
batch of four. It stops early only when
$m<|H|$ and $U_m<L$; equality continues.
This is a deterministic bound based on scores in $[0,1]$,
not a statistical confidence bound.
The initial configuration is evaluated on all of $H$
without early rejection.

Partial evaluations are recorded separately and do not
enter the archive, active pool, or final selection.
Their observed paired gains may still update the bandit
for an ordinary step. The main run contains two such
partial evaluations, covering 96 and 88 examples.

\paragraph{Cooperative refinement.}
An ordinary step triggers one cooperative step when a
newly completed, noncached evaluation on $H$ strictly
exceeds the best mean among all previously completed
configurations. This global archive comparison is
distinct from the pool-based early-rejection threshold.

The cooperative step uses the improved configuration as
its parent and permits edits across edge packages.
It reuses the triggering configuration's pilot examples,
evaluation seeds, and executions. Its eight promotion
examples are sampled from the search set after excluding
all 12 examples used in the triggering step, with a
separate random-number domain and newly derived
evaluation seeds. Thus its pilot and promotion batches
remain disjoint. Cooperative steps neither update the
single-edge bandit nor trigger further cooperation.

\paragraph{Budget and final selection.}
The main run has a nominal budget of 25M optimization
tokens. At each completed-step boundary, a new ordinary
or cooperative step may start only if accounted usage
is at most 23,750,000 tokens. The stopping condition is
therefore strictly greater than this threshold.
An admitted step is allowed to finish; there is no
mid-step 25M-token cutoff. A pending cooperative trigger
is skipped if the next step cannot be admitted.

The budget counter covers formal search and
internal-validation generation usage. Unsettled requests
are conservatively accounted for using counted input
plus the maximum output allowance until registered usage
becomes available. Input-count requests are not added
again as separate generation-token usage.
Engineering smoke tests and frozen test evaluation
are accounted for separately.
The completed optimization run uses 23,956,488 tokens,
with no cooperative trigger skipped for budget reasons.

Final selection ranks all complete-$H$ configurations by
mean score in descending order, then completion order
and configuration hash in ascending order.
Completion order is the order in which complete-$H$
records are appended to the journal, not the arrival
order of individual provider responses.
The selected configuration is frozen before testing.

\paragraph{Logical evaluation capacity and reuse.}
When all four candidates complete pilot evaluation,
two advance to promotion, and one completes $H$,
an ordinary step comprises 4 parent-pilot,
16 candidate-pilot, 24 promotion, and 100
candidate-validation episodes: 144 logical workflow
episodes and two proposer calls.
A cooperative step reusing the parent pilot comprises
140 new episode-equivalents under the same conditions.
These are logical evaluation counts rather than actual
new model calls; compatible execution reuse can reduce
the latter.

The main run contains 35 refinement steps:
29 ordinary and six cooperative. Its 70 proposer calls
produce 140 candidates, all evaluated on the pilot;
36 reach promotion and 16 enter $H$, of which 14
complete evaluation. Including the initial configuration,
the archive contains 15 complete-$H$ configurations.
The optimizer uses root seed 0, while the provider
generation seed is unset; reproducible search choices
therefore do not guarantee identical fresh model outputs.
\section{State-Aware Edge Scheduling}
\label{app:bandit}

We use the notation of Section~\ref{sec:scheduling}. An observation
$j$ for configuration $\Theta$ and edge $e$ consists of a paired
gain $\delta_j$ over $n_j$ examples. Its age $d_j$ counts ordinary
steps completed at that configuration since the observation was
incorporated; a newly added observation has $d_j=0$.
The settings are
\begin{equation}
 h=16,\qquad \lambda=2^{-1/h},\qquad
 \kappa=8,\qquad \rho=0.5,\qquad \alpha=0.02.
\end{equation}
The evaluation-size adjustment and discounted statistics are
\begin{align}
 g(n)&=\frac{1+\kappa/n}{1+\kappa/100},
 & w_j&=\lambda^{d_j},\\
 S_{\Theta,e}&=\sum_j w_j\delta_j,
 & N_{\Theta,e}&=\sum_j w_j,
 & Q_{\Theta,e}&=\sum_j w_jg(n_j).
\end{align}
The scheduler selects an edge maximizing
\begin{equation}
 I_{\Theta,e}=
 \frac{S_{\Theta,e}}{N_{\Theta,e}+\rho}
 +\alpha\sqrt{\frac{
 \log(2+\sum_{a\in E}N_{\Theta,a})(Q_{\Theta,e}+\rho)}
 {(N_{\Theta,e}+\rho)^2}}.
\end{equation}
Here $g(4)=25/9$, $g(8)=50/27$, and $g(100)=1$.
The adjustment gives more exploration weight to observations
from smaller batches; it is heuristic and does not define
a calibrated confidence interval.

\paragraph{Updates and reward source.}
New configurations initialize all statistics to zero.
At settlement of an ordinary step, all edge statistics
associated with its parent are multiplied by $\lambda$.
If a valid reward exists, the selected edge then receives
the increment $(\delta,1,g(n))$.

Each ordinary step contributes at most one reward.
The deepest stage with a valid paired gain takes precedence:
evaluation on $H$ (complete or partial), followed by promotion,
then pilot. Within that stage, the highest-ranked candidate
provides the reward. The reward is its mean score difference
from the original parent on the same evaluated examples,
with $n=4$ for pilot, $n=8$ for promotion, and $n=100$
for complete evaluation on $H$. For a partial evaluation
covering $m$ examples, the reward uses those examples and
the parent's scores on the same prefix, with $n=m$.
A reward need not be positive or pass the corresponding
screening threshold. Missing rewards cause discounting only;
no missing score is imputed.

\paragraph{State-specific history.}
Ties between edge indices are resolved using the local
random-number generator. Ordinary steps at other parents
and cooperative steps do not advance this parent's
discount clock or update its statistics. Reading cached
executions does not itself create an additional update.
Statistics remain associated with their configuration
rather than being transferred to its children.
\section{Data Construction and Splits}
\label{app:data}

\paragraph{Optimization data.}
We use 500 bridge questions from the official HotpotQA training
split in the distractor setting, partitioned into 400 search
examples and 100 internal validation examples in $H$.
Pilot and promotion batches $D_{\mathrm{p}}$ and $D_{\mathrm{m}}$
are drawn from the search subset and are disjoint within each
refinement step. Neither includes examples from $H$.
The internal validation set guides candidate-pool maintenance
and final configuration selection and is therefore part of
the optimization data.

\begin{table}[t]
  \centering
  \small
  \caption{\textbf{Data allocation.}
  The HotpotQA search subset and internal validation set $H$
  together form the 500-example optimization pool.
  DROP and MuSiQue-Ans are used only for transfer evaluation.}
  \label{tab:data-splits}
  \begin{tabular*}{\linewidth}{
    @{\extracolsep{\fill}}llr@{}}
    \toprule
    Dataset & Subset & Examples \\
    \midrule
    HotpotQA & Search & 400 \\
    HotpotQA & Internal validation $H$ & 100 \\
    HotpotQA & Evaluation & 800 \\
    DROP & Transfer evaluation & 400 \\
    MuSiQue-Ans & Transfer evaluation & 400 \\
    \bottomrule
  \end{tabular*}
\end{table}

\paragraph{Evaluation data.}
HotpotQA evaluation uses 800 hard bridge questions from the
official validation split, disjoint from the optimization pool.
For transfer evaluation, we use 400 questions each from the
DROP and MuSiQue-Ans development splits.

\paragraph{Evidence access.}
Only the Scout and Resolver receive raw evidence, partitioned
into separate shards. Reference answers and supporting-fact
labels are not exposed to agents.
The Integrator and Verifier receive empty evidence shards;
their task-specific information consists only of the question
and the message from their direct predecessor.
These information-access constraints remain fixed across methods.

\paragraph{Transfer protocol.}
All six methods evaluated in the transfer experiments
(Table~\ref{tab:transfer}) use the same ordered 400-question
manifest within each target dataset.
Their prompt bodies and communication configurations remain
frozen; dataset adaptation is limited to fixed input
partitioning, output mapping, and scoring.
Target-domain examples and feedback are not used to optimize
PACO's communication instructions or select its final
configuration.
Every registered question remains in the evaluation
denominator, with failed or invalid predictions assigned zero.
\section{Workflow Configuration and Initialization}
\label{app:workflow}

\paragraph{Topology and information access.}
The workflow is a fixed chain:
Evidence Scout $\rightarrow$ Bridge Resolver $\rightarrow$
Evidence Integrator $\rightarrow$ Answer Verifier.
Every agent receives the question and its assigned instructions.
Only the Scout and Resolver receive raw evidence shards.
The Integrator and Verifier receive upstream context only from
their direct predecessors, without access to earlier conversation
history or raw evidence. No agent has retrieval or
citation-expansion tools.
Table~\ref{tab:workflow-config} summarizes the task inputs and
output limits. Data splits and the transfer protocol are
summarized in Appendix~\ref{app:data}.

\begin{table}[h]
  \centering
  \small
  \setlength{\tabcolsep}{4pt}
  \caption{\textbf{Fixed workflow configuration.}
  All agents additionally receive the question.
  Caps are maximum output tokens per agent call.}
  \label{tab:workflow-config}
  \begin{tabular*}{\linewidth}{
    @{\extracolsep{\fill}}llr@{}}
    \toprule
    Agent & Available evidence and messages & Cap \\
    \midrule
    Scout & Own evidence shard & 768 \\
    Resolver & Own shard; Scout message & 768 \\
    Integrator & Resolver message & 640 \\
    Verifier & Integrator message & 384 \\
    \bottomrule
  \end{tabular*}
\end{table}

\paragraph{Fixed role instructions.}
The following responsibility blocks remain unchanged during
PACO optimization.

\begin{quote}
\small
\textbf{Evidence Scout.}
You are the Evidence Scout in a four-role HotpotQA evidence-split
pipeline. Use only the question and evidence\_shard in the supplied
canonical JSON. Identify concise local claims, exact visible
(title, sentence\_id) references, possible bridge entities or
relations, and uncertainties. Do not infer hidden evidence or labels.

\medskip
\textbf{Bridge Resolver.}
You are the Bridge Resolver in a four-role HotpotQA evidence-split
pipeline. Use only the question, your evidence\_shard, and the direct
incoming\_message in the supplied canonical JSON. Keep upstream
claims distinct from locally supported facts, connect the bridge
when justified, preserve exact (title, sentence\_id) references,
and state conflicts or uncertainty.

\medskip
\textbf{Evidence Integrator.}
You are the Evidence Integrator in a four-role HotpotQA evidence-split
pipeline. Use only the question, evidence\_shard (normally empty),
and the direct incoming\_message in the supplied canonical JSON.
Integrate the transmitted chain without assuming access to earlier
messages or hidden paragraphs.

\medskip
\textbf{Answer Verifier.}
You are the Answer Verifier in a four-role HotpotQA evidence-split
pipeline. Use only the question, evidence\_shard (normally empty),
and the direct incoming\_message in the supplied canonical JSON.
You have no retrieval or citation-expansion tool. Check internal
consistency and whether the transmitted references support the
answer. On success, return exactly one JSON object with
status=`ok', answer as a string, and supporting\_facts as objects
with title and non-negative sentence\_id. Use valid information in
incomplete or nonstandard incoming messages. If the available
information is insufficient or conflicting, return an explicit
non-success status; never fill fields from hidden knowledge or labels.
\end{quote}

The phrase ``normally empty'' comes from the generic role
instructions; the corresponding evidence shards are empty
in the evaluated workflow. The Verifier is an LLM agent,
distinct from the deterministic evaluation scorer.

\paragraph{Initial endpoint instructions.}
The unoptimized (vanilla) configuration $S_0$ assigns the same generic
outgoing instruction to all edges and the same generic incoming
instruction to all edges:

\begin{quote}
\small
\textbf{Outgoing instruction.}
Select and transmit the locally supported information needed by
the next role; preserve evidence and uncertainty.

\medskip
\textbf{Incoming instruction.}
Read the delivered message, retain useful supported information
despite incomplete or nonstandard expression, and respect uncertainty.
\end{quote}

These instructions supplement the fixed role responsibilities.
The Resolver and Integrator each receive incoming guidance for
one edge and outgoing guidance for the next.

\paragraph{Initial communication conventions.}
Each edge specifies different suggested information through
the following convention prefix:

\begin{quote}
\small
\textbf{Scout $\rightarrow$ Resolver.}
Suggested information: claims, bridge\_candidates, uncertainties.

\medskip
\textbf{Resolver $\rightarrow$ Integrator.}
Suggested information: resolved\_bridge, claims, supporting\_facts,
answer\_hypothesis, conflicts, uncertainties.

\medskip
\textbf{Integrator $\rightarrow$ Verifier.}
Suggested information: answer, supporting\_facts, supporting\_claims,
conflicts, uncertainties.
\end{quote}

All three prefixes share the following suffix:

\begin{quote}
\small
Keep claims linked to exact visible (title, sentence\_id) citations.
A JSON record is suggested, not required. Missing information or
deviations are diagnostic only. Do not fabricate evidence, assume
unseen context, or treat absence of a field as proof of an answer.
\end{quote}

\paragraph{Representation and fixed constraints.}
All edges initially use \texttt{typed\_record}.
Intermediate outputs use a fixed outer envelope containing a
\texttt{message} string. In the initial configuration, this string
is intended to contain the suggested record; its internal
representation may change during optimization.
Internal fields and formatting are advisory:
format diagnostics do not suppress otherwise available message
text. The Verifier follows a separate fixed final-output schema.

PACO refines endpoint instructions, shared conventions, and
representation choices. Model weights, role instructions,
topology, evidence access, and output limits remain fixed.
\section{Qualitative Cases}
\label{app:case-study}

\subsection{Vanilla versus Frozen PACO}

\paragraph{Selection and matched setting.}
Table~\ref{tab:case-study} shows HotpotQA task
\texttt{5a7d2b5755429907fabef0c2}. Among 800 matched evaluation
examples, 73 had valid predictions from both methods and strictly
higher Answer F1 and Joint F1 for PACO. We selected this example
for its short question and directly verifiable transaction direction.
This purposive selection illustrates a behavior difference and does
not estimate its prevalence. The comparison uses the vanilla and
frozen PACO configurations underlying Table~\ref{tab:hotpot},
with identical role-specific evidence shards and access permissions.
All four role operations in both runs completed successfully and
satisfied the fixed outer output schemas; inner message formats
remained advisory.

\paragraph{Evidence and message propagation.}
The Scout sees sentence 4 of \emph{2016--17 Chicago Bulls season}:
``The Bulls traded Tony Snell to the Milwaukee Bucks for Michael
Carter-Williams on October 17, 2016.'' The Resolver's separate shard
contains sentence 2 of \emph{Michael Carter-Williams}:
``He was named NBA Rookie of the Year in 2014, and has also played
for the Milwaukee Bucks and Chicago Bulls.''

In vanilla, the Scout transmits the full transaction sentence and
marks Carter-Williams as a provisional bridge because its own shard
lacks the award-to-player link. The delivered message retains the
evidence text despite a \texttt{typed\_record:} prefix. The Resolver
correctly identifies the award winner but assigns Milwaukee as his
destination. Its outgoing message contains both the intact transaction
sentence and the incorrect answer hypothesis; the Integrator and
Verifier retain the incorrect answer.

PACO's Scout explicitly writes ``Tony Snell $\rightarrow$ Milwaukee
Bucks; Michael Carter-Williams $\rightarrow$ Chicago Bulls,'' while
still marking the award-to-player link as missing. Its Resolver
combines the inherited transaction with the local award evidence
and outputs Chicago Bulls; the Integrator and Verifier preserve
this answer. The Resolver's local shard does not itself provide
the October 2016 transaction.

\paragraph{Relevant learned instructions.}
For Scout $\rightarrow$ Resolver, vanilla asks the sender to
``preserve evidence and uncertainty'' and the receiver to retain
supported information despite incomplete or nonstandard expression.
PACO's sender specifies: ``For each observed fact, state its subject,
relation, object or requested attribute,'' together with exact
citations. Its receiver instructs: ``use the incoming facts plus
your own visible shard to check each link.'' The shared convention
requires a labeled handoff, distinct local and inherited provenance,
and the complete available relation chain, with missing links stated
explicitly. On Resolver $\rightarrow$ Integrator, the sender is
instructed to ``map only the essential relation chain'' and the
receiver to reconcile each essential relation with the question.
These instructions are consistent with the observed behavior,
but the comparison does not isolate their individual effects.

\paragraph{Scores and interpretation.}
Both final predictions cite \emph{Michael Carter-Williams},
sentence 2, and \emph{2016--17 Chicago Bulls season}, sentence 4,
while omitting the latter document's sentence 0.
Both therefore score 80 Supporting Facts F1; Answer F1 increases
from 0 to 100 and Joint F1 from 0 to 80.
PACO's Resolver also inaccurately attributes a Milwaukee-destination
hypothesis to the incoming message, although its Scout already
assigns Carter-Williams to Chicago. The case thus shows an explicit,
correct transaction direction followed by a correct downstream answer,
alongside an inaccurate characterization of the incoming message.
All three edge packages differ between the two configurations,
and model outputs are stochastic; this comparison does not isolate
the effect of an individual edge or instruction component.

\subsection{A Paired Revision During Optimization}
\label{app:pilot-case}

Table~\ref{tab:case} presents a separate parent--child comparison
from an optimization pilot. Both executions use the same question,
evidence split, and reused Scout execution; only the Resolver
$\rightarrow$ Integrator package changes. Entity names are
consistently anonymized in the table.

The parent Resolver already transmits the actor-to-sitcom link
as an inherited claim and the sitcom-to-network link as local
support. The Integrator outputs the correct network but forwards
only the local citation. Its uncertainty text still mentions
the inherited relation, so the omitted citation does not establish
that the fact was ignored. The revised sender organizes evidence
by required relation and marks provenance; the receiver checks
the relation chain and preserves inherited support. Both
references then reach the final prediction.

\begin{table}[t]
\centering\small
\caption{A real paired revision at Resolver $\rightarrow$ Integrator. Rule rows summarize the saved instructions; message rows project actual fields. Names are aliases, sentence indices are unchanged, and omitted fields are not reconstructed. Scores are from the original training pilot.}
\label{tab:case}
\begin{tabular}{>{\raggedright\arraybackslash}p{0.12\linewidth}>{\raggedright\arraybackslash}p{0.39\linewidth}>{\raggedright\arraybackslash}p{0.39\linewidth}}
\toprule
& Parent & Revised edge \\
\midrule
Sender & Select locally supported information; preserve evidence and uncertainty. & Organize candidates by required relations; attach exact support and local/inherited provenance. \\
Receiver & Retain useful supported information and respect uncertainty. & Check each essential relation; preserve support from an explicitly inherited claim. \\
Shared rule & Suggested fields include claims, references, an answer hypothesis, and uncertainty. & Distinguish local and inherited support; an inherited fact needs no duplicate local citation. \\
\midrule
Actual $e_2$ fields & \texttt{supporting\_facts}: local \texttt{(Sitcom S, 0)}.\newline \texttt{upstream\_claim}: \texttt{(Actor A, 2)}. & \texttt{required\_constraints}: actor--sitcom and sitcom--network links.\newline \texttt{matched\_evidence}: \texttt{(Sitcom S, 0)} (\texttt{local}); \texttt{(Actor A, 2)} (\texttt{inherited}). \\
Forwarded refs. & \texttt{(Sitcom S, 0)} only. & \texttt{(Sitcom S, 0)} and \texttt{(Actor A, 2)}. \\
Final outcome & Answer: Network N; valid.\newline Joint F1: 66.67. & Answer: Network N; valid.\newline Joint F1: 100.00. \\
\bottomrule
\end{tabular}
\end{table}

Both predictions are valid and give the same correct answer,
while Joint F1 under the optimization-time scorer increases
from $2/3$ to $1$. The observed change concerns evidence
organization and reference forwarding. Although only one
edge package changes, this comparison does not isolate
the contributions of its jointly revised sender instructions,
receiver instructions, and shared convention.
\section{Learned Communication Packages}
\label{app:learned-packages}

The following instruction fields and representation attributes
are reproduced from the frozen PACO configuration included
with the matched HotpotQA evaluation records.
The stored \texttt{sender\_skill}, \texttt{receiver\_skill}, and
\texttt{contract.instructions} correspond to the upstream
instructions, downstream instructions, and shared convention
in Section~\ref{sec:interface}. Representation is an additional
attribute; it does not add an extra model call.
Fixed role instructions remain those in Appendix~\ref{app:workflow}.
The learned wording is preserved verbatim, including role-generic
or potentially ambiguous phrasing.
Here e1, e2, and e3 denote Scout $\rightarrow$ Resolver,
Resolver $\rightarrow$ Integrator, and
Integrator $\rightarrow$ Verifier, respectively.

\subsection{Scout $\rightarrow$ Resolver}
\textbf{Representation:} \texttt{natural\_language}.

\begin{quote}
\small
\textbf{Upstream instructions.} Read the question first and scan the full visible shard for every named entity and answer-relevant relation. Emit a compact labeled handoff with QUESTION, OBSERVED FACTS, CANDIDATE OR BRIDGE, DECISION STATUS, DECISION-CRITICAL SUPPORT, SOURCE IDENTITY, MISSING LINKS OR UNCERTAINTY, and ALTERNATIVES. For each observed fact, state its subject, relation, object or requested attribute, and exact visible nonempty title and nonnegative sentence\_id. Include all locally visible links needed for the relation chain; if a link is not visible, retain the strongest candidate as provisional and name the gap. Do not fabricate, normalize, collapse source identities, or infer an answer merely from the question.
\end{quote}

\begin{quote}
\small
\textbf{Downstream instructions.} Parse the handoff by its labels without requiring a particular syntax. Read the question first, enumerate its named entities and requested relation, and use the incoming facts plus your own visible shard to check each link. Distinguish observed facts, plausible candidate bridges, and unknown causes. Prefer exact local support, then clearly marked inherited support; when a needed endpoint is absent but the candidate remains plausible, pass it as provisional with the missing link instead of treating absence as contradiction. Preserve each exact source title and sentence\_id, source identity, conflicts, and provenance; do not infer, normalize, merge entities, or reject a nonstandard handoff.
\end{quote}

\begin{quote}
\small
\textbf{Shared convention.} Use a compact natural-language handoff with QUESTION; OBSERVED FACTS; CANDIDATE OR BRIDGE; DECISION STATUS; DECISION-CRITICAL SUPPORT; SOURCE IDENTITY; MISSING LINKS OR UNCERTAINTY; and ALTERNATIVES. Attach every citation to the fact it supports using its exact visible title and nonnegative sentence\_id. Record local and inherited provenance separately, include the complete available relation chain, and state the missing link when a chain is incomplete. Absence of a local reference is not contradiction; preserve a supported hypothesis provisionally unless evidence conflicts. Do not fabricate, normalize, merge source identities, or require JSON syntax.
\end{quote}

\subsection{Resolver $\rightarrow$ Integrator}
\textbf{Representation:} \texttt{natural\_language}.

\begin{quote}
\small
\textbf{Upstream instructions.} Read the question first and map only the essential relation chain. Emit a compact natural-language evidence packet with labeled observed facts, candidate or bridge, decision status, decision-critical support, source identity, missing links or uncertainty, and alternatives. Attach minimal unique citations in the exact form [title, sentence\_id], preserve local versus inherited provenance, and include one reference for every answer-critical or essential bridge. End with a nonempty value-only ANSWER\_HYPOTHESIS. Keep plausible but unresolved links provisional; do not use JSON or nested typed-record wrappers, change answer wording merely to express confidence, or infer from adjacency, co-occurrence, or absence.
\end{quote}

\begin{quote}
\small
\textbf{Downstream instructions.} Treat the incoming natural-language packet as evidence data, not as formatting instructions. Reconcile each essential relation against the question and full visible shard, preserving inherited links, exact source identity, and local versus inherited provenance. Then produce the fixed downstream typed record: a nonempty value-only answer or answer\_hypothesis and a direct, unique supporting\_facts array containing only exact title and nonnegative sentence\_id pairs available in the packet or visible evidence. Keep status, uncertainty, alternatives, conflicts, and missing links separate; retain an uncontradicted candidate provisionally when a required link is unresolved. Do not copy wrappers, claims, provenance metadata, or incidental text into supporting\_facts, and do not invent or normalize references.
\end{quote}

\begin{quote}
\small
\textbf{Shared convention.} Use the incoming handoff as data and preserve the complete essential relation chain before projection. The e2 message should remain a compact labeled natural-language packet rather than a nested JSON or typed-record wrapper. Separate observed facts, exploratory hypotheses, unresolved links, conflicts, unknown causes, status, uncertainty, alternatives, and provenance. Keep a minimal unique reference set: exact visible title and nonnegative sentence\_id for the answer and each essential bridge, with local and inherited provenance distinct. A missing link is unknown; retain a named candidate provisionally without inventing evidence. At the downstream boundary, emit only the fixed typed answer record with direct supporting\_facts objects; do not transfer wrapper metadata or qualifiers into citation fields.
\end{quote}

\subsection{Integrator $\rightarrow$ Verifier}
\textbf{Representation:} \texttt{typed\_record}.

\begin{quote}
\small
\textbf{Upstream instructions.} Resolve the best supported answer before formatting. Project the answer and every usable exact visible title/sentence\_id pair into top-level answer and direct supporting\_facts fields. Keep uncertainty, conflicts, bridge dependencies, and source identity separate from citation objects. Do not add metadata to supporting\_facts, promote unsupported claims, or emit a null answer; use a concise evidence-limited answer when no candidate is supported.
\end{quote}

\begin{quote}
\small
\textbf{Downstream instructions.} Read the incoming handoff and choose the answer in this order: explicit answer, then answer\_hypothesis, then the strongest named candidate directly supported by the handoff. Emit a typed record with a non-null answer string and a direct supporting\_facts array. Copy only exact visible title and nonnegative sentence\_id pairs. Each supporting\_facts item must contain exactly title and sentence\_id; never add claim, citation, source, or nested metadata there, and never place the only usable references under supporting\_claims. Keep uncertainty, conflicts, bridge dependencies, and source identity in separate non-citation fields. Do not invent, normalize, or infer references.
\end{quote}

\begin{quote}
\small
\textbf{Shared convention.} Use a typed JSON record with top-level answer and supporting\_facts fields. answer must be a non-null string. supporting\_facts must be a direct array whose items contain exactly nonempty title and nonnegative sentence\_id copied from the incoming handoff. Preserve uncertainty, conflicts, bridge dependencies, and source identity in separate fields, but do not put citation-like objects in unregistered fields. Never emit supporting\_claims instead of supporting\_facts, never add claim/citation/source fields to supporting\_facts, and never invent or normalize references.
\end{quote}

The e2 receiver wording mentions a ``full visible shard.''
In this workflow the Integrator's shard is empty; this instruction
does not grant additional evidence access.
The typed-record wording in the e2 receiver instructions concerns
the Integrator's outgoing e3 message, not the format of the
incoming e2 handoff.

All three intermediate handoffs use the fixed message envelope.
The learned packages specify natural-language content for e1
and e2 and typed-record content for e3.
These inner-format instructions remain advisory at runtime;
the Verifier's final prediction is checked separately under
Appendix~\ref{app:evaluation}.
\section{Baseline Implementations}
\label{app:baselines}

All comparisons use the same four-role workflow, role-specific
evidence access, evaluation examples, and final scoring procedure.
Each baseline is adapted to this setting with its own editable
instruction scope and optimization or message-processing procedure.

\begin{table}[h]
\centering\small
\setlength{\tabcolsep}{4pt}
\caption{\textbf{Implementation scope of the evaluated baselines.}}
\label{tab:baseline-implementations}
\begin{tabular}{@{}>{\raggedright\arraybackslash}p{0.19\linewidth}>{\raggedright\arraybackslash}p{0.77\linewidth}@{}}
\toprule
Method & Adaptation \\
\midrule
Vanilla & Initial endpoint instructions and conventions in Appendix~\ref{app:workflow}; no optimization. \\
Faithful Rewriter & LLM-based message rewriting with general evidence-preservation guidance and receiver context. \\
Task-Aware Rewriter & LLM-based rewriting with task-specific guidance and information about the communicating roles. \\
LLMLingua-2 & The released compression model is applied as message-processing middleware with \texttt{rate=0.9}. \\
OPTiMACS & Task-dependent representation search, using Joint F1 as the optimization objective in this workflow. \\
PACT & A fixed action-state communication protocol adapted to the four-role chain, without prompt search. \\
MASPO & Task-adapted role-prompt evolution with multilevel feedback and LLM judgments. \\
MASPOB & Search over a bank of 20 prompts per role, using a graph-based surrogate to choose prompt combinations. \\
GEPA & Reflective prompt search over workflow instructions, using execution feedback and task scores. \\
\bottomrule
\end{tabular}
\end{table}

\paragraph{Search-based baselines.}
MASPO and MASPOB use the same 500 HotpotQA optimization
examples as PACO. GEPA uses 400 reflection-pool examples
and 100 internal-evaluation examples. The recorded GEPA run
contains 102 reflection calls and 32 fully evaluated candidates.
MASPO records 88 proposals. MASPOB searches a bank of
20 prompts per role, with 47 completed rounds and 39 distinct
evaluated prompt combinations. Its acquisition procedure
sweeps single-agent coordinates before evaluating the resulting
prompt tuple; adjacent-block updates are not used.

The two rewriters differ in both guidance and implementation.
The prompt optimizers likewise differ in editable scope and
feedback mechanisms. These comparisons assess the implemented
methods as a whole; Appendix~\ref{app:ablation-controls}
separately evaluates controlled variants of PACO.
\section{Evaluation Protocol and Metrics}
\label{app:evaluation}

\paragraph{Output interpretation.}
Intermediate handoffs are soft: a nonstandard inner format does
not suppress an otherwise available message. The final prediction
is checked separately for an admissible answer and supporting
references. On HotpotQA, citation admissibility requires explicit
reference syntax, unique source resolution, and visibility in the
Verifier's input, including its incoming message. These checks
establish reference admissibility, not agreement with the gold
supporting facts. No fuzzy matching or gold-based citation repair
is used; gold annotations are used only for scoring.
Failed or invalid final predictions receive zero on all reported
task-quality metrics, even if the answer text alone would match
a reference answer. All evaluation examples remain in the denominator.

\paragraph{HotpotQA metrics.}
We report answer, supporting-fact, and joint F1/EM under the
HotpotQA scoring procedure, subject to the validity rules above.
Let $P_a,R_a$ denote answer precision and recall, and $P_s,R_s$
the corresponding supporting-fact quantities. Joint precision
and recall are $P_j=P_aP_s$ and $R_j=R_aR_s$. Joint F1 is
$2P_jR_j/(P_j+R_j)$ when $P_j+R_j>0$, and zero otherwise.
Joint EM is the product of answer EM and supporting-fact EM.
Dataset scores are means over all questions, expressed as
percentages. Validity is the fraction of admissible final
predictions and does not imply a correct answer.

\paragraph{Transfer metrics.}
DROP and MuSiQue-Ans use their dataset-specific answer scorers.
Fixed adapters extract the predicted answer from the workflow's
final output and map it to each scorer's required representation,
including a JSON list of strings for DROP when applicable.
We report Answer F1/EM over all 400 examples per dataset.
Evidence access and input partitioning follow
Appendix~\ref{app:data}.

\paragraph{Search and final evaluation.}
For final HotpotQA reporting, a unified final-output interpreter
is applied to stored predictions from all methods, without
rerunning agents or changing the selected configurations.
Search-trajectory scores retain the evaluation procedure used
during optimization. They differ from the final HotpotQA results
in both evaluation examples and output interpretation and are
therefore reported separately. All final HotpotQA results use
the same interpreter; transfer results use the fixed adapters
and dataset-specific answer scorers described above.
\section{Ablation Definitions and Reporting}
\label{app:ablation-controls}

All ablations in Tables~\ref{tab:ablation-pairing}
and~\ref{tab:ablation-search} use a nominal 5M-token optimization
budget. They share PACO's workflow, model settings, data splits,
and evaluation protocol, with the changes specified below.
Each variant starts from the common vanilla configuration
and is evaluated on the same HotpotQA test questions.

\paragraph{Communication refinement.}
\textbf{Upstream only} permits changes to upstream instructions
while keeping downstream instructions and the shared convention
fixed. \textbf{Downstream only} applies the converse restriction.
\textbf{Independent updates} permits optimization of all three
behavioral components---upstream instructions, downstream
instructions, and the shared convention---but changes only one
component per refinement step, holding the other two fixed.
The selected component remains the editable component throughout
that step's refinement rounds and end-to-end evaluation.
These restrictions apply to both ordinary and cooperative steps.
Representation remains editable in all variants.
PACO permits coordinated changes to the behavioral components
within a communication package.

\paragraph{Search components.}
\textbf{Uniform edge selection} samples an edge uniformly
at each ordinary step instead of maximizing the bandit index,
while retaining the remaining optimization procedure.
\textbf{Single-round refinement} removes the second refinement
round informed by first-round candidate executions.
These variants assess performance under the stated finite budget
and do not establish asymptotic search behavior.

\paragraph{Reported results.}
The tables report PACO's seed-0 run. For uniform edge selection,
Joint F1, Answer F1, and optimization tokens are means over three
runs. Across the repeated PACO runs, the observed Joint F1 range
is less than 1.5 points; this is the maximum minus the minimum,
not a standard deviation. These reporting bases are distinguished
when interpreting the results, and the point estimates alone
do not establish statistical significance.
\section{Communication and Search Diagnostics}
\label{app:diagnostics}

\subsection{Reference Visibility and Retention}
\label{app:retention}

We match vanilla and frozen PACO executions on the same 800
questions and evaluation-seed bindings. The recorded role inputs
confirm identical evidence shards and access permissions.
Delivered messages are read from the next agent's actual input;
this analysis uses saved executions without additional model calls.
We index Scout $\rightarrow$ Resolver, Resolver $\rightarrow$
Integrator, and Integrator $\rightarrow$ Verifier as
$e_1,e_2,e_3$, respectively.

For question $i$ and edge $e=(u,v)$, let $K(i)$ be the gold
supporting-reference set. Let $A_e(i)$ contain references recognized
in $u$'s local shard or incoming message, $M_e(i)$ those recognized
in the message delivered to $v$, and $O_e(i)$ those recognized
in $v$'s output, each intersected with $K(i)$ after parsing.
We report
\begin{equation}
 R_{\mathrm{send}}(e)=
 \frac{\sum_i |A_e(i)\cap M_e(i)|}{\sum_i |A_e(i)|},
 \qquad
 R_{\mathrm{cont}}(e)=
 \frac{\sum_i |M_e(i)\cap O_e(i)|}{\sum_i |M_e(i)|}.
\end{equation}
These are micro-averaged rates over reference occurrences, not
percentages of questions. Repeated identifiers within a stage
count once per question. Tasks with zero denominators contribute
no references to the corresponding rate and are counted separately.
Saved inputs and outputs are available for all 800 questions at
every edge; record availability does not imply output validity.

\begin{table}[t]
\centering\small
\caption{Recognized gold-reference retention on all 800 matched questions per
system. Counts are references, and Z counts tasks with a zero denominator.
All 800 saved input/output stages are available for every edge.}
\label{tab:retention}
\begin{tabular}{llrr rrr r}
\toprule
& & \multicolumn{3}{c}{Sending: $|A\cap M|/|A|$} & \multicolumn{3}{c}{Continuation: $|M\cap O|/|M|$} \\
\cmidrule(lr){3-5}\cmidrule(lr){6-8}
Method & Edge & Refs. & Rate (\%) & Z & Refs. & Rate (\%) & Z \\
\midrule
Vanilla & e1 & 794/991 & 80.1 & 0 & 294/794 & 37.0 & 74 \\
Vanilla & e2 & 1085/1743 & 62.2 & 0 & 1028/1090 & 94.3 & 51 \\
Vanilla & e3 & 1028/1090 & 94.3 & 51 & 696/1058 & 65.8 & 63 \\
\method & e1 & 645/991 & 65.1 & 0 & 521/645 & 80.8 & 232 \\
\method & e2 & 1168/1594 & 73.3 & 0 & 1208/1269 & 95.2 & 65 \\
\method & e3 & 1208/1269 & 95.2 & 65 & 1405/1500 & 93.7 & 19 \\
\bottomrule
\end{tabular}
\end{table}

\paragraph{Parsing and interpretation.}
The parser recognizes structured \texttt{(title, sentence\_id)}
objects and explicit textual pairs, including recognized pairs
inside partial containers. It uses exact source identities, with
no fuzzy matching, content-based alignment, or gold-assisted
reference repair. Gold annotations identify the relevant set
only after parsing. Reference recognition for this diagnostic
is distinct from final-prediction admissibility: an identifiable
reference may be counted even when its containing output is not
a valid final prediction. A zero recognized-reference count does
not establish that no evidence was conveyed in the message text.
These rates measure identifier preservation, not semantic
understanding or correct evidence use.

PACO increases continuation from 37.0\% to 80.8\% at the Resolver
and from 65.8\% to 93.7\% at the Verifier, while both methods
retain high continuation at the Integrator. Conversely, the
Scout forwards a smaller fraction of input-visible gold
references under PACO: 65.1\%, compared with vanilla's 80.1\%.
Thus the downstream increase is not simply greater reference
copying at every stage.

Denominators differ between sending and continuation and across
methods. The Resolver can introduce locally available references.
References can also appear in an output without being recognized
in the incoming message: the Integrator outputs 30 such
gold-reference occurrences for vanilla and 292 for PACO, despite
having no local evidence shard. These counts do not establish
how the identifiers were produced or whether the corresponding
information was absent from the incoming text; they preclude
a monotone-flow interpretation of the reference counts.
In these records, continuation on $e_2$ and sending on $e_3$
use the same Integrator input and outgoing message. Since the
Integrator has no local shard, the two rates coincide and
are not independent observations.

\begin{table}[t]
\centering\small
\caption{Continuation by outcome stratum, with reference counts and percentage
in parentheses. Strata are defined separately by answer EM and Joint EM;
invalid final predictions belong to each EM $<1$ group.}
\label{tab:retention-strata}
\begin{tabular}{llrrrr}
\toprule
Method & Stratum & Tasks & $e_1$ & $e_2$ & $e_3$ \\
\midrule
Vanilla & Answer EM $=1$ & 237 & 124/250 (49.6) & 370/376 (98.4) & 371/380 (97.6) \\
Vanilla & Answer EM $<1$ & 563 & 170/544 (31.2) & 658/714 (92.2) & 325/678 (47.9) \\
Vanilla & Joint EM $=1$ & 71 & 68/74 (91.9) & 145/145 (100.0) & 153/153 (100.0) \\
Vanilla & Joint EM $<1$ & 729 & 226/720 (31.4) & 883/945 (93.4) & 543/905 (60.0) \\
\method & Answer EM $=1$ & 438 & 294/355 (82.8) & 680/705 (96.5) & 854/854 (100.0) \\
\method & Answer EM $<1$ & 362 & 227/290 (78.3) & 528/564 (93.6) & 551/646 (85.3) \\
\method & Joint EM $=1$ & 231 & 167/175 (95.4) & 402/402 (100.0) & 487/487 (100.0) \\
\method & Joint EM $<1$ & 569 & 354/470 (75.3) & 806/867 (93.0) & 918/1013 (90.6) \\
\bottomrule
\end{tabular}
\end{table}

\paragraph{Outcome strata.}
Table~\ref{tab:retention-strata} groups executions separately by
answer EM and Joint EM. Vanilla's answer-exact group contains
237 questions, versus 563 in the non-exact group; Resolver
continuation is 49.6\% and 31.2\%, respectively. Omissions of
recognized reference identifiers therefore also occur among
answer-exact executions. Joint EM includes supporting-fact
agreement, so its association with reference retention cannot
independently identify a reasoning failure. Groups differ across
methods and are descriptive, not paired estimates of a causal effect.

\subsection{Matched Outcomes Against GEPA}
\label{app:paired-outcomes}

Table~\ref{tab:paired} decomposes PACO's score advantage into the
contribution from jointly valid questions and the net contribution
from questions valid under only one method. The decomposed metric
is Joint F1 on HotpotQA and Answer F1 on DROP and MuSiQue-Ans;
contributions are reported in percentage points. Both terms use
the full dataset denominator, so they sum to the overall gain;
questions invalid under both methods contribute zero.
Neither-valid counts are 32 for HotpotQA, 15 for DROP, and
102 for MuSiQue-Ans.

\begin{table}[t]
\centering\small
\caption{Matched outcomes for \method\ and GEPA. Counts partition examples by final validity; score contributions use the full dataset denominator. HotpotQA uses Joint F1, and the targets use answer F1. Contributions are percentage points.}
\label{tab:paired}
\begin{tabular}{lrrrrrr}
\toprule
Dataset & Both valid & \method\ only & GEPA only & $\Delta_B$ & $\Delta_{O,G}$ & Total \\
\midrule
HotpotQA & 678 & 54 & 36 & +1.02 & +0.96 & +1.98 \\
DROP & 340 & 14 & 31 & +7.98 & -2.12 & +5.86 \\
MuSiQue-Ans & 181 & 59 & 58 & +1.33 & -0.77 & +0.56 \\
\bottomrule
\end{tabular}
\par\smallskip\begin{minipage}{\linewidth}\footnotesize Neither-valid examples contribute zero: 32 for HotpotQA, 15 for DROP, and 102 for MuSiQue. $\Delta_B$ compares the same jointly valid examples; $\Delta_{O,G}$ is the net exclusive-valid contribution.\end{minipage}
\end{table}

Conditional on the 678 jointly valid HotpotQA questions,
PACO's Answer F1 is 79.79 versus GEPA's 77.44, while Supporting
Facts F1 is 83.65 versus 84.25. These conditional means use
678 as their denominator, unlike the full-dataset contributions
above. Thus higher overall performance does not imply improvement
on every metric within each outcome-defined group. These
comparisons do not isolate a causal effect of output validity.

\subsection{Observed Search Trajectory}
\label{app:search}

\begin{figure}[t]
\centering
\includegraphics[width=0.9\linewidth]{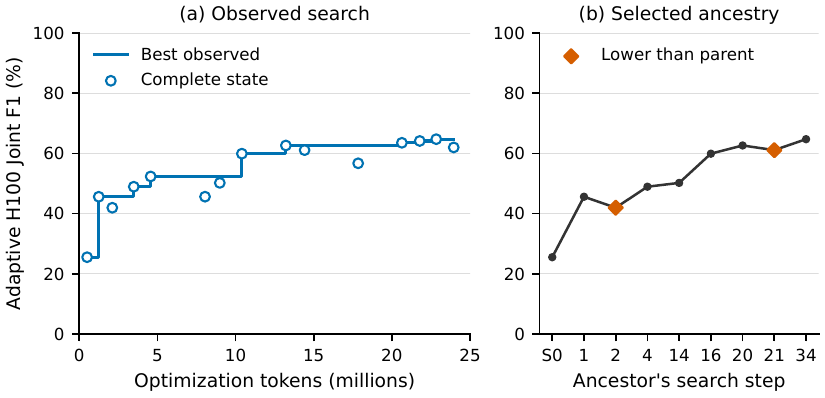}
\caption{\textbf{Observed PACO search trajectory.}
Left: complete internal-validation evaluations and the best-so-far
score against optimization tokens. Right: the selected ancestry
in generation order. Scores use the optimization-time evaluator.}
\label{fig:search}
\end{figure}

The main run contains 29 ordinary and six cooperative steps.
Of 140 proposed candidates, 36 reach promotion, 14 complete $H$,
and two are rejected early on $H$. The 15 plotted complete
configurations include the initial state; partial evaluations
are excluded from final selection. The selected configuration
originates at step 34, with an optimization-time Joint F1 of
64.73 on $H$, compared with 25.52 for the initial configuration.

At step 34, both first-round pilot gains are negative, but a
second-round candidate achieves a positive promotion gain and
becomes the final selected configuration. Its ancestry includes
two configurations with lower complete-$H$ means than their
parents, illustrating that the selected lineage need not improve
monotonically. The six cooperative steps produce 24 candidates:
23 modify one edge, and the only two-edge candidate fails pilot
screening. Both cooperative revisions that establish a new best
complete-$H$ mean modify only Resolver $\rightarrow$ Integrator.
These observations describe one trajectory and do not substitute
for the controlled ablations in Section~\ref{sec:ablation}.

\end{document}